\pdfoutput=1
\documentclass[11pt]{article}

\usepackage{acl}

\usepackage{times}
\usepackage{relsize}
\usepackage{latexsym}
\usepackage{graphicx}
\usepackage{todonotes}

\usepackage{listings}
\usepackage{color}

\definecolor{dkgreen}{rgb}{0,0.6,0}
\definecolor{gray}{rgb}{0.5,0.5,0.5}
\definecolor{mauve}{rgb}{0.58,0,0.82}

\usepackage[T1]{fontenc}
\usepackage[utf8]{inputenc}

\usepackage{microtype}

\usepackage{fdsymbol}

\DeclareCaptionStyle{ruled}{labelfont=normalfont,labelsep=colon,strut=off} % DO NOT CHANGE THIS
\title{Solving Probability and Statistics Problems by Program Synthesis}

\author{Leonard Tang \\
  Harvard University\\
  Mathematics\\\And
  Elizabeth Ke \\
  MIT\\
  Mathematics\\\And
  Nikhil Singh \\
  MIT\\
  Media Lab\\\AND
  Nakul Verma \\
  Columbia University\\
  Computer Science Department\\\And
  Iddo Drori \\
  MIT\\
  EECS\\}

\begin{document}
\maketitle
\begin{abstract}
We solve university level probability and statistics questions by program synthesis using OpenAI's Codex, a Transformer trained on text and fine-tuned on code. We transform course problems from MIT's 18.05 Introduction to Probability and Statistics and Harvard's STAT110 Probability into programming tasks. We then execute the generated code to get a solution. Since these course questions are grounded in probability, we often aim to have Codex generate probabilistic programs that simulate a large number of probabilistic dependencies to compute its solution. Our approach requires prompt engineering to transform the question from its original form to an explicit, tractable form that results in a correct program and solution. To estimate the amount of work needed to translate an original question into its tractable form, we measure the similarity between original and transformed questions. Our work is the first to introduce a new dataset of university-level probability and statistics problems and solve these problems in a scalable fashion using the program synthesis capabilities of large language models.
\end{abstract}

\section{Introduction}
Let's say we play a game where I keep flipping a coin until I get heads. If the first time I get heads is on the $n$-th coin, then I pay you $2n - 1$ dollars. How much would you pay me to play this game?

How would one solve this problem in an automated fashion?
Existing approaches to solving such a problem, typical in university level probability and statistics courses, overwhelmingly hinge upon directing foundation models to formulate answers in a \textit{deductive} fashion, whether via a sequence of steps \cite{hendrycksmath2021} or formal operations \cite{amini-etal-2019-mathqa}. 

An alternate compelling approach is to simulate a given task on a large scale and aggregate results across multiple scenarios. Such an approach, usually dubbed as probabilistic programming in the literature \cite{pmlr-v15-wingate11a}, offers a flexible mechanism for solving a variety of probabilistic tasks.

\begin{figure*}[!h]
    \centering
    \includegraphics[width=\textwidth]{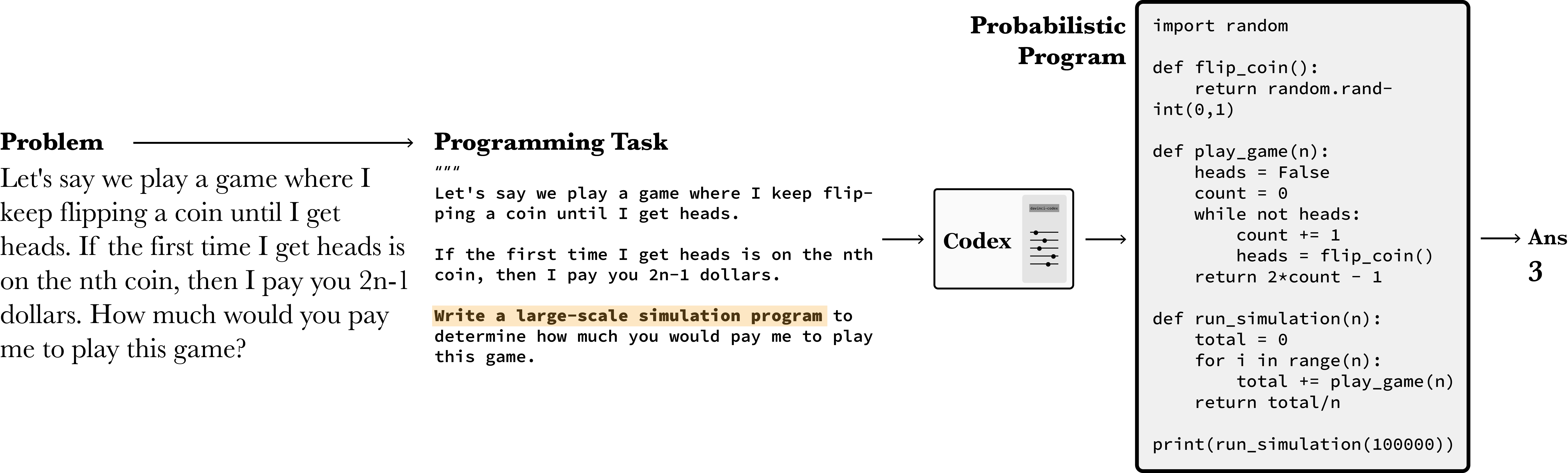}
    \vspace{0.05in}
    \caption{Probabilistic Simulation Program Workflow Example: (i) The original problem is translated into a programming task that asks Codex to simulate a large number of probabilistic scenarios, (ii) Codex generates such a program, and (iii) the program is executed to yield an answer.}
    \label{fig:prob-workflow}
\end{figure*}

Inspired by this insight, our goal is to solve probability problems both via simulation and direct methods by leveraging the power of a program synthesizer such as OpenAI's Codex \cite{chen2021evaluating}. Codex is a Transformer model trained on text and fine-tuned on code, which has the capacity to write programs that can simulate arbitrary stochastic tasks. Our core approach is neatly demonstrated in Figure \ref{fig:prob-workflow}. We transform raw question text into a programming task. This is then fed into Codex, generating a probabilistic program. We can then execute this code to get the correct response.

To the best of our knowledge, we are the first to propose such a  simulation based approach to solve probability questions. To evaluate the efficacy of our approach, we collect two sets of 20 undergraduate-level probability and statistics problems, curated from MIT's 18.05 and Harvard's STAT110.

The key to our success lies in the carefully engineered prompts we present to Codex. Critically, we introduce the notion of \textit{Concept-Grounded Task} prompting, i.e.\ priming Codex with related concepts and problem-solving strategies in its prompt (cf.\ Figure \ref{fig:concept-workflow}).

\subsection{Related Work}

\paragraph{Foundation models.}
Foundation models \cite{bommasani2021opportunities} such as GPT-3 \cite{NEURIPS2020_1457c0d6} have demonstrated impressive and unforeseen emergent capabilities from their learning process, including aptitude in automatic speech recognition, vision, commonsense reasoning, and more \cite{9054476, dosovitskiy2020vit, Bosselut2019COMETCT}. For the task of answering questions specifically, such models have recently achieved strong performance \cite{rajpurkar-etal-2018-know}. However, when tasked with solving university-level quantitative problems, foundation model performance is poor \cite{hendrycksmath2021}.

\paragraph{Probability benchmarks.} 
Though recent works have introduced datasets, such as MATH, MAWPS, MathQA, Math23k, and GSM8K \cite{hendrycksmath2021, koncel-kedziorski-etal-2016-mawps, amini-etal-2019-mathqa, wang-etal-2017-deep, cobbe2021training}, that focus on benchmarking mathematical question answering, including probability questions, but all of these works only consider grade-school level question difficulty. We are the first to present two datasets of undergraduate-level probability and statistics questions.

\section{Dataset}
We introduce two datasets of questions from two separate undergraduate-level probability and statistics courses of varying difficulty and a set of quantitative finance interview questions. We describe the datasets below:

\begin{enumerate}
    \item The first dataset consists of applied questions in probability. We take 20 questions that have numerical answers from MIT's 18.05: Introduction to Probability and Statistics \cite{18_05_ocw}. Course topics include various probability and statistics concepts, including counting, conditional probability, discrete and continuous random variables, expectation and variance, central limit theorem, joint distributions, maximum likelihood estimators, Bayesian updating, null hypothesis significance testing, and confidence intervals.
    \item The second dataset, in contrast to the first dataset, consists of conceptual questions in probability and statistics. We take 20 questions that have numerical answers from Harvard's STAT110: Probability \cite{stat110home} and Brainstellar \cite{brainstellar} online catalogue of quantitative finance probability brainteasers. Topics include: distributions, moment generating functions, expectation, variance, covariance, correlation, conditional probability, joint distributions, marginal distributions, conditional distributions, limit theorems, and Markov chains.
\end{enumerate}

See Appendix for the list of all questions in our datasets.

\begin{figure*}[t!]
    \centering
    \includegraphics[width=\textwidth]{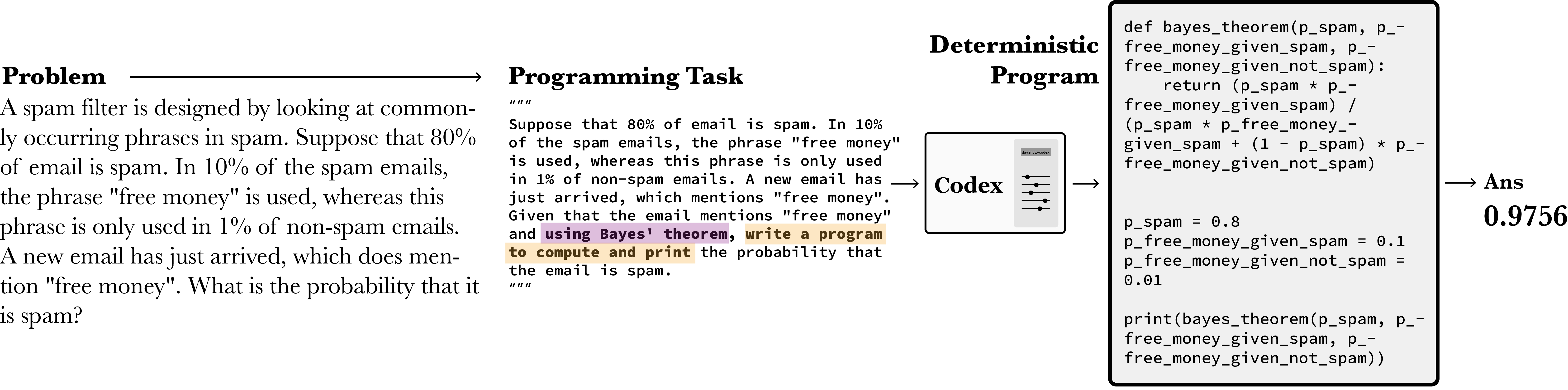}
    \vspace{0.05in}
    \caption{Concept-Grounded Workflow Example: (i) The original problem is translated into a programming task that includes Bayes' Theorem within its context, (ii) Codex generates a program, and (iii) the program is executed to yield an answer.}
    \label{fig:concept-workflow}
\end{figure*}

\section{Methods}

\subsection{Models and Evaluation}
As our core program synthesizer, we leverage OpenAI's Codex \cite{chen2021evaluating}.  Given a raw question text, we use the following experimentation pipeline: we convert each question into a programming task, prompt Codex with the task to get a programmatic solution, and execute the program generated by Codex, comparing the execution result to the ground truth solution. 

A critical component of this workflow is \textit{prompt engineering} (i.e.\ the conversion from question to programming task). We examine prompt engineering in further detail in Section \ref{lab:prompt_egg} and measure the degree to which we manipulate the prompt in Section \ref{lab:measure_transform}.

\subsection{Prompt Engineering}
\label{lab:prompt_egg}
Large generative language models, including Codex, are known to be extremely sensitive to input prompts \cite{DBLP:journals/corr/abs-2102-07350}. Below, we outline and describe three classes of prompts that we communicate to Codex with and their associated effects:

\begin{itemize}
    \item \textbf{Program Task Specification:} One class of prompts converts probability questions into direct task specifications \cite{DBLP:journals/corr/abs-2102-07350}. In this case, these specifications are explicit programming assignments. For instance, if the original question is "What is the probability of flipping two heads in a row given a fair coin?", the corresponding task specification would be "Write a program that computes the probability of flipping two heads in a row given a fair coin."  While these prompts occasionally suffice, in many instances additional prompt manipulation is required.
     \item \textbf{Probabilistic Simulation Programming}: While the above classes of prompts produce programs deterministic in nature, our third prompting technique hinges upon the power of probabilistic simulation programs, i.e. programs that simulate a large number of scenarios and aggregate results across simulations to determine an approximate answer. To trigger such simulation behavior in Codex, we include in our prompt the substring "Write a large-scale simulation program to estimate," followed by the desired task. Figure \ref{fig:prob-workflow} presents an example using such a prompting scheme.
    \item \textbf{Concept-Grounded Task}: An extremely useful extension beyond \textit{Program Task Specification} is to include relevant information pertaining to both the question and program contexts. The question context includes related topics or mathematical rules to use. This is primarily represented in the form of canonical equations, definitions, and theorems. Providing Codex with hints on problem-solving \textit{strategy} involving these concepts is extremely helpful. For instance, if a question is related to the concept of Bayes' Theorem, appending the transformed prompt with an explicit instruction to use Bayes' Theorem results in a correct program. This is demonstrated in Figure \ref{fig:concept-workflow}. In addition to the question context, it may be useful to specify the programming context, including which packages or libraries the program will load and use, such as packages for symbolic math, integration, or optimization.
\end{itemize}

\section{Results}

\subsection{Evaluating Program Output}
\begin{figure}[ht!]
    \centering
    \includegraphics[width=\columnwidth]{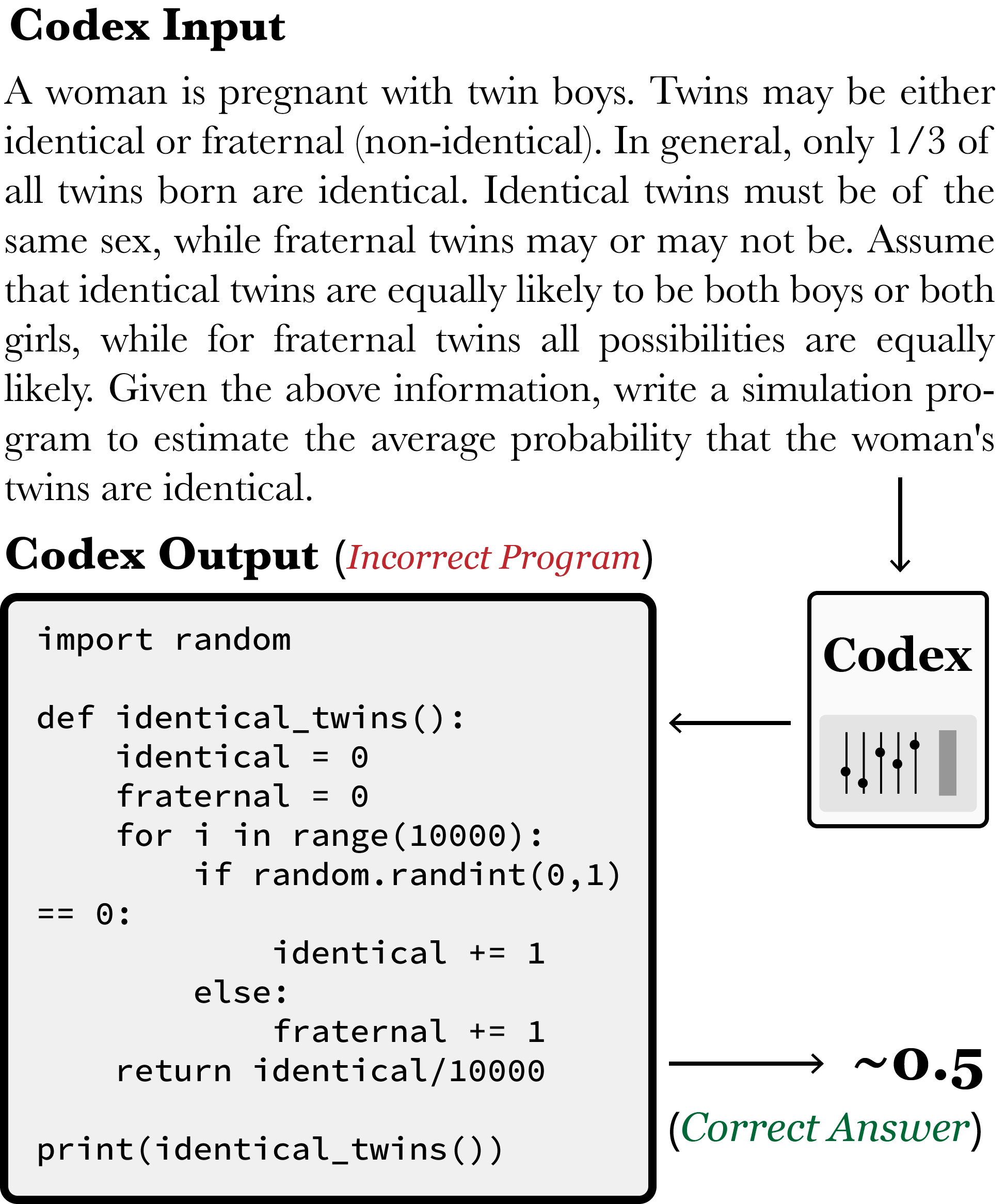}
    \caption{Evaluating Codex requires more than just checking the numerical answer. It requires evaluating the logic of the program, and seeing that it actually answers the program.
    Though the computed answer matches the ground truth solution of 0.5, the Codex program is written with the intention of calculating the unconditional probability of having fraternal twins (assuming fraternal and identical twins have equal probabilities), and completely ignores the conditioning information in the problem.}
    \label{fig:wrong_logic}
\end{figure}

It is possible for Codex to generate code which appears to yield a correct answer, but in reality is not a correct method to solve the problem. Thus, we explicitly inspect the code generated by Codex to check for its logical correctness. See Figure \ref{fig:wrong_logic} for an explicit example which we encountered.

Another peculiar challenge in evaluating Codex's generated simulation programs lies in the approximate nature of their output. Specifically, we can only achieve perfect accuracy in the limit of simulation scale. Hence, we designate a numerical output resulting from program execution as correct when it is within $1\%$ of the ground truth solution.

\subsection{Achieving Perfect Results}
\label{lab:measure_transform}

Following the methods discussed in Section \ref{lab:prompt_egg}, we are able to generate correct programs for all questions in both datasets. See the Appendix for additional detail regarding the original question text, the corresponding prompt-engineered transformation, the generated program, and finally the program evaluation for each question in each dataset.

Since our approach involves prompt engineering in the process of translating a question to a programming task, we seek a concrete measure of the effort necessary in this transformation. Our metric is computed as follows: for any given pair of original question and transformed programming task, we compute the cosine similarity between the Sentence-BERT \cite{reimers-gurevych-2019-sentence} embedding of the question and Sentence-BERT embedding of the task. 

\begin{figure}[ht!]
    \centering
    \includegraphics[width=\columnwidth]{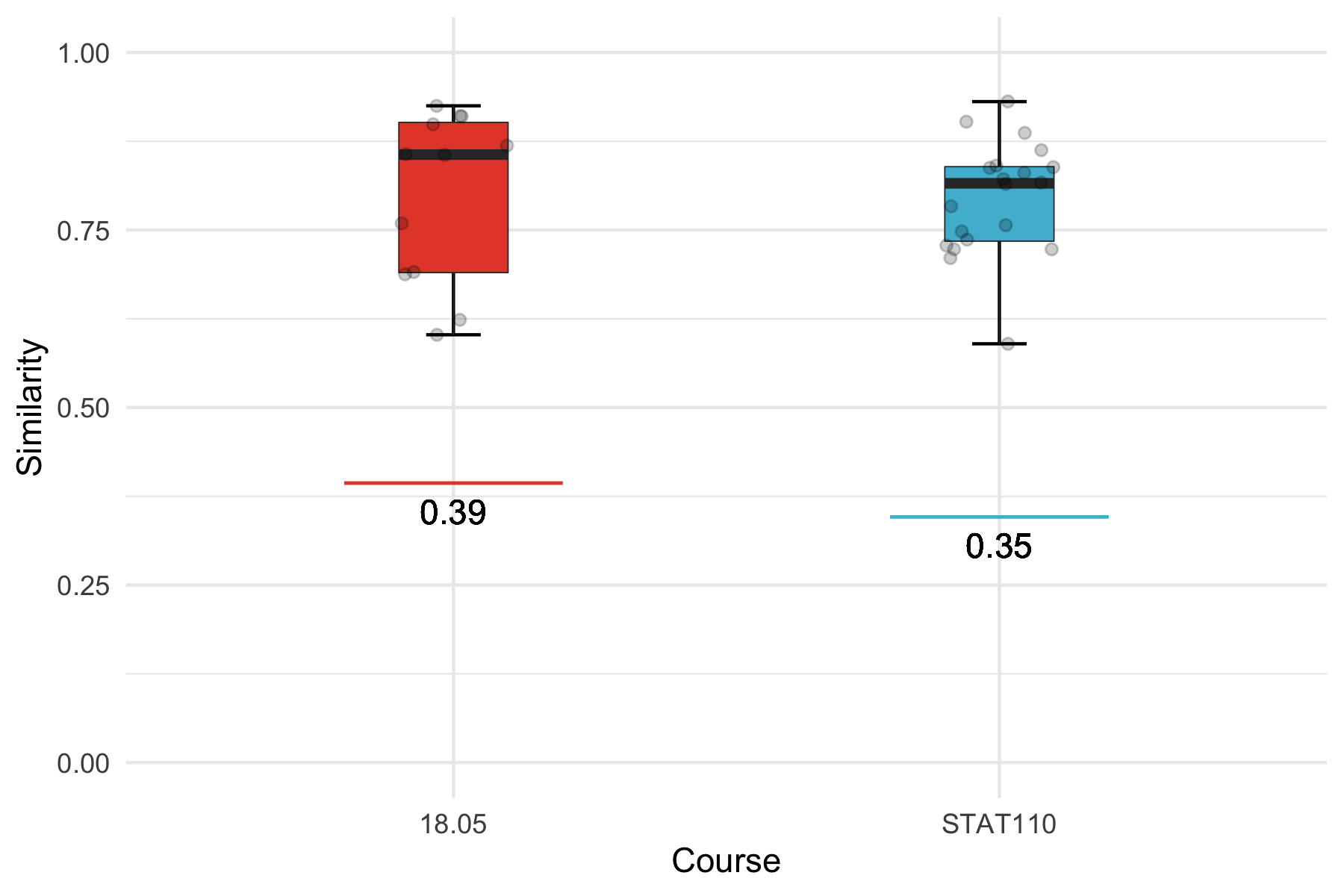}
    \caption{Sentence-BERT Similarity between original questions and programming tasks by course. Each course's transformation similarities are contextualized by baseline question similarities, i.e. the average pairwise similarity between the original questions.}
    \label{fig:course_similarity}
\end{figure}

Figure \ref{fig:course_similarity} shows that we have an average similarity of 0.80 in MIT's 18.05 and an average similarity of 0.79 in STAT110. As a baseline reference we also include the average pairwise similarity score among the original questions, thus indicating we only need minor changes to the text.

Finally, since theses course are available online, we verify that Codex is not overfitting to training data, by writing and solving our own novel questions.

\subsection{Implementation Details}
We fix all Codex's hyperparameters to be the same for all experiments: \textit{top-p}\footnote{\textit{top-p} designates the portion $p$ of the token probability mass a language model samples from at each step.} is set to $1$, sampling temperature is set to $0$ (i.e.\ argmax), and maximum sequence length is set to $750$ tokens. Both frequency and presence penalty are set to $0$, and we do not halt on any stop sequences. We use the \textit{davinci} engine for all of our generations.

\section{Conclusion}
To the best of our knowledge, this is the first work to present a state-of-the-art method that leverages the program synthesis and probabilistic simulation capabilities of foundation models to solve university-level probability and statistics problems. Through the use of prompt engineering, including priming Codex with concepts and problem-solving strategies, we achieve full correctness on our novel datasets.
We plan to expand our work to further understand the underlying structure of programming tasks that are amenable to Codex manipulation, and in a similar vein move towards the automatic translation of questions to programming tasks.

\newpage
\clearpage

\bibliography{bibliography}
\bibliographystyle{acl_natbib}

\onecolumn
\appendix
\section{MIT 18.05: Introduction to Probability and Statistics}

\begin{table*}[h]
\small \centering
    \begin{tabular}{|p{3.5cm}|p{11.5cm}|}
        \hline
        Topic & Probability\\
        \hline
        Original question & In class we worked with non-transitive dice: Red: 3 3 3 3 3 6; Green: 1 4 4 4 4 4; White: 2 2 2 5 5 5. Find the probability that white beats green.\\
        \hline
        Codex input & Given non-transitive dice: Red: 3 3 3 3 3 6; Green: 1 4 4 4 4 4; White: 2 2 2 5 5 5, write a program that computes the probability that white beats green.\\
        \hline
        Codex output & \begin{lstlisting}
import itertools
    
red = [3, 3, 3, 3, 3, 6]
green = [1, 4, 4, 4, 4, 4]
white = [2, 2, 2, 5, 5, 5]

def prob_white_beats_green(red, green, white): 
    rolls = list(itertools.product(red, green, white))
    wins = [1 for (r, g, w) in rolls if w > g]
    return sum(wins) / len(rolls)

prob_white_beats_green(red, green, white) 
        \end{lstlisting}
        \\
        \hline
        Solution & 7/12\\
        \hline
    \end{tabular}
\caption*{Question 1 from 18.05}
\end{table*}

\begin{table*}[h]
\small \centering
    \begin{tabular}{|p{3.5cm}|p{11.5cm}|}
        \hline
        Topic & Probability\\
        \hline
        Original question & Calculate the probability of getting a two-pair poker hand.\\
        \hline
        Codex input & A hand is a set of 5 cards that are drawn randomly from a standard 52 card deck with 13 ranks of 4 cards each. \newline \newline A two-pair poker hand is a hand that contains 3 unique ranks, where no more than 2 cards in the hand can share the same rank. That is, 3 or more cards cannot share the same rank. \newline \newline Write a program that generates simulations for calculating the average probability of getting a two-pair poker hand.\\
        \hline
        Codex output & \begin{lstlisting}
import random

def get_rank_counts(hand):
    """"""
    Returns a dictionary with ranks as keys and the number of cards with that rank as values.
    """"""
    rank_counts = {}
    for card in hand:
        rank = card[0]
        if rank in rank_counts:
            rank_counts[rank] += 1
        else:
            rank_counts[rank] = 1
    return rank_counts

def has_two_pair(hand):
    """"""
    Returns True if the hand has two pairs, and False otherwise.
    """"""
    rank_counts = get_rank_counts(hand)
    two_pair = False
    for rank in rank_counts:
        if rank_counts[rank] == 2:
            if two_pair:
                return True
            else:
                two_pair = True
    return False

def run_simulation():
    """"""
    Runs a simulation of drawing 5 cards randomly from a deck of 52 cards.
    """"""
    two_pair = 0
    for i in range(1000000):
        deck = [r+s for r in '23456789TJQKA' for s in 'SHDC']
        random.shuffle(deck)
        hand = deck[:5]
        if has_two_pair(hand):
            two_pair += 1
    return two_pair/1000000

print(run_simulation())
        \end{lstlisting}\\
        \hline
        Solution & 0.047539\\
        \hline
    \end{tabular}
\caption*{Question 2 from 18.05}
\end{table*}

\begin{table*}[h]
\small \centering
    \begin{tabular}{|p{3.5cm}|p{11.5cm}|}
        \hline
        Topic & Probability\\
        \hline
        Original question & Calculate the probability of getting a three-of-a-kind poker hand.\\
        \hline
        Codex input & A hand is a set of 5 cards that are drawn randomly from a standard 52 card deck with 13 ranks of 4 cards each. \newline \newline A three-pair poker hand is a hand that contains exactly 3 cards that share one rank, 1 card of another rank, and a final card of another rank. \newline \newline Write a program that generates simulations for calculating the average probability of getting a three-pair poker hand.\\
        \hline
        Codex output & \begin{lstlisting}
import random

def get_rank_counts(hand):
    """
    Return a dictionary with ranks as keys and their counts as values.
    """
    ranks = [card[0] for card in hand]
    return {rank: ranks.count(rank) for rank in ranks}

def is_three_pair(hand):
    """
    Return True if hand is a three pair hand.
    """
    rank_counts = get_rank_counts(hand)
    return 3 in rank_counts.values()

def get_three_pair_probability(n):
    """
    Return the probability of getting a three pair hand.
    """
    three_pair_count = 0
    for _ in range(n):
        deck = [
            '2C', '2D', '2H', '2S',
            '3C', '3D', '3H', '3S',
            '4C', '4D', '4H', '4S',
            '5C', '5D', '5H', '5S',
            '6C', '6D', '6H', '6S',
            '7C', '7D', '7H', '7S',
            '8C', '8D', '8H', '8S',
            '9C', '9D', '9H', '9S',
            'TC', 'TD', 'TH', 'TS',
            'JC', 'JD', 'JH', 'JS',
            'QC', 'QD', 'QH', 'QS',
            'KC', 'KD', 'KH', 'KS',
            'AC', 'AD', 'AH', 'AS'
        ]
        hand = random.sample(deck, 5)
        if is_three_pair(hand):
            three_pair_count += 1
    return three_pair_count / n

print(get_three_pair_probability(1000000))
        \end{lstlisting}\\
        \hline
        Solution & 0.021128\\
        \hline
    \end{tabular}
\caption*{Question 3 from 18.05}
\end{table*}

\begin{table*}[h]
\small \centering
    \begin{tabular}{|p{3.5cm}|p{11.5cm}|}
        \hline
        Topic & Probability\\
        \hline
        Original question & In class we worked with non-transitive dice: Red: 3 3 3 3 3 6; Green: 1 4 4 4 4 4; White: 2 2 2 5 5 5. Find the probability that green beats red.\\
        \hline
        Codex input & Given two non-transitive dice: \newline Red: 3 3 3 3 3 6; \newline Green: 1 4 4 4 4 4. \newline Write a program that computes the probability that green beats red.\\
        \hline
        Codex output & \begin{lstlisting}
def dice_prob(dice1, dice2):
    '''
    dice1: list of dice 1
    dice2: list of dice 2
    '''
    # Your code here
    red_wins = 0
    green_wins = 0
    for i in dice1:
        for j in dice2:
            if i > j:
                red_wins += 1
            elif j > i:
                green_wins += 1
    return green_wins / (red_wins + green_wins)

# Test Cases
dice1 = [3, 3, 3, 3, 3, 6]
dice2 = [1, 4, 4, 4, 4, 4]
print(dice_prob(dice1, dice2))
        \end{lstlisting}\\
        \hline
        Solution & 25/36\\
        \hline
    \end{tabular}
\caption*{Question 4 from 18.05}
\end{table*}

\begin{table*}[h]
\small \centering
    \begin{tabular}{|p{3.5cm}|p{11.5cm}|}
        \hline
        Topic & Conditional Probability\\
        \hline
        Original question & Mr. Smith has two children. At least one of them is a boy. What is the probability that both children are boys?\\
        \hline
        Codex input & Write a program that computes the probability that both children are boys given that one of them is a boy. \\
        \hline
        Codex output & \begin{lstlisting}
# probability that one child is a boy
p_one_boy = 1/4

# probability that both children are boys
p_both_boys = 1/4

# probability that one of them is a boy
p_one_is_boy = 3/4

# probability that both children are boys given that one of them is a boy
p_both_boys_given_one_boy = p_both_boys / p_one_is_boy

print(p_both_boys_given_one_boy)
        \end{lstlisting}\\
        \hline
        Solution & 1/3\\
        \hline
    \end{tabular}
\caption*{Question 5 from 18.05}
\end{table*}

\begin{table*}[h]
\small \centering
    \begin{tabular}{|p{3.5cm}|p{11.5cm}|}
        \hline
        Topic & Discrete Random Variables, Standard Deviation\\
        \hline
        Original question & Let X be the result of rolling a fair 4-sided die. Let Y be the result of rolling a fair 6-sided die. Let Z be the average of X and Y. Find the standard deviation of X.\\
        \hline
        Codex input & In Probability, \newline Write a program that computes the variance and standard deviation of rolling a fair 4-sided die 1000000 times.\\
        \hline
        Codex output & \begin{lstlisting}
import random

def variance(lst):
    mean = sum(lst) / len(lst)
    return sum((i - mean) ** 2 for i in lst) / len(lst)

def standard_deviation(lst):
    return variance(lst) ** 0.5

def main():
    lst = []
    for i in range(1000000):
        lst.append(random.randint(1, 4))
    print(variance(lst))
    print(standard_deviation(lst))

if __name__ == '__main__':
    main()
        \end{lstlisting}\\
        \hline
        Solution & 1.118\\
        \hline
    \end{tabular}
\caption*{Question 6 from 18.05}
\end{table*}

\begin{table*}[h]
\small \centering
    \begin{tabular}{|p{3.5cm}|p{11.5cm}|}
        \hline
        Topic & Discrete Random Variables, Standard Deviation\\
        \hline
        Original question & Let X be the result of rolling a fair 4-sided die. Let Y be the result of rolling a fair 6-sided die. Let Z be the average of X and Y. Find the standard deviation of Y.\\
        \hline
        Codex input & In Probability, Write a program that computes the variance and standard deviation of rolling a fair 6-sided die 1000000 times.\\
        \hline
        Codex output & \begin{lstlisting}
import statistics

def variance_and_standard_deviation():
    data = []
    for i in range(1000000):
        data.append(random.randint(1,6))
    print(statistics.variance(data))
    print(statistics.stdev(data))

variance_and_standard_deviation()
        \end{lstlisting}\\
        \hline
        Solution & 1.708\\
        \hline
    \end{tabular}
\caption*{Question 7 from 18.05}
\end{table*}

\begin{table*}[h]
\small \centering
    \begin{tabular}{|p{3.5cm}|p{11.5cm}|}
        \hline
        Topic & Discrete Random Variables, Standard Deviation\\
        \hline
        Original question & Let X be the result of rolling a fair 4-sided die. Let Y be the result of rolling a fair 6-sided die. Let Z be the average of X and Y. Find the standard deviation of Z.\\
        \hline
        Codex input & Let X be the result of rolling a fair 4-sided die. Let Y be the result of rolling a fair 6-sided die. Let Z be the average of X and Y. Write a program that computes the standard deviation of Z. \\
        \hline
        Codex output & \begin{lstlisting}
import math

def standard_deviation(lst):
    mean = sum(lst) / len(lst)
    return math.sqrt(sum([(x - mean) ** 2 for x in lst]) / len(lst))

def main():
    X = [1, 2, 3, 4]
    Y = [1, 2, 3, 4, 5, 6]
    Z = []
    for i in X:
        for j in Y:
            Z.append((i + j) / 2)
    print(standard_deviation(Z))

if __name__ == "__main__":
    main()
        \end{lstlisting}\\
        \hline
        Solution & 1.021\\
        \hline
    \end{tabular}
\caption*{Question 8 from 18.05}
\end{table*}

\begin{table*}[h]
\small \centering
    \begin{tabular}{|p{3.5cm}|p{11.5cm}|}
        \hline
        Topic & Discrete Random Variables, Expectation\\
        \hline
        Original question & Let X be the result of rolling a fair 4-sided die. Let Y be the result of rolling a fair 6-sided die. You win 2X dollars if X>Y and lose 1 dollar otherwise. After playing this game 60 times, what is your expected total gain?\\
        \hline
        Codex input & In Probability, \newline X is the result of rolling a fair 4-sided die and Y is the result of rolling a fair 6-sided die. You win 2X if X > Y and lose 1 otherwise. \newline Write a program that computes the expected total gain after rolling both dice 60 times.\\
        \hline
        Codex output & \begin{lstlisting}
import random

def roll_dice(): 
    return random.randint(1, 4)

def roll_dice_6():
    return random.randint(1, 6)

def play_game():
    total_gain = 0
    for i in range(60):
        x = roll_dice()
        y = roll_dice_6()
        if x > y:
            total_gain += 2 * x
        else:
            total_gain -= 1
    return total_gain

def main():
    total_gain = 0
    for i in range(1000):
        total_gain += play_game()
    print(total_gain / 1000)

if __name__ == '__main__':
    main()
        \end{lstlisting}\\
        \hline
        Solution & 55\\
        \hline
    \end{tabular}
\caption*{Question 9 from 18.05}
\end{table*}

\begin{table*}[h]
\small \centering
    \begin{tabular}{|p{3.5cm}|p{11.5cm}|}
        \hline
        Topic & Counting, Continuous Variables\\
        \hline
        Original question & Boxes of Raisin Bran cereal are 30cm tall. Due to settling, boxes have a higher density of raisins at the bottom (h = 0) than at the top (h = 30). Suppose the density (in raisins per cm of height) is given by f(h) = 40 - h. How many raisins are in a box?\\
        \hline
        Codex input & Boxes of Raisin Bran cereal are 30cm tall. \newline Suppose the density of raisins per cm of height is given by f(h) = 40 - h. \newline Using scipy.integrate, write a program that computes how many raisins are in a box.\\
        \hline
        Codex output & \begin{lstlisting}
import numpy as np
import scipy.integrate as integrate

def f(h):
    return 40 - h

def integrate_f(a, b):
    return integrate.quad(f, a, b)

print(integrate_f(0, 30))
        \end{lstlisting}\\
        \hline
        Solution & 750\\
        \hline
    \end{tabular}
\caption*{Question 10 from 18.05}
\end{table*}

\begin{table*}[h]
\small \centering
    \begin{tabular}{|p{3.5cm}|p{11.5cm}|}
        \hline
        Topic & Probability, Continuous Variables\\
        \hline
        Original question & Boxes of Raisin Bran cereal are 30cm tall. Due to settling, boxes have a higher density of raisins at the bottom (h = 0) than at the top (h = 30). Suppose the density (in raisins per cm of height) is given by f(h) = 40 - h. What is the probability that a random raisin is in the bottom third of the box?\\
        \hline
        Codex input & h ranges from 0 to 30. Integrate f(h) = 40 - h from 0 to 10. Divide by the integration of h from 0 to 30.\\
        \hline
        Codex output & \begin{lstlisting}
def f(h):
    return 40 - h

def integrate(f, a, b, N):
    """
    Integrates f from a to b using N intervals
    """
    dx = (b - a) / N
    total = 0
    for i in range(N):
        total += f(a + i * dx)
    return total * dx

print(integrate(f, 0, 10, 100000) / integrate(f, 0, 30, 100000))
        \end{lstlisting}\\
        \hline
        Solution & 7/15\\
        \hline
    \end{tabular}
\caption*{Question 11 from 18.05}
\end{table*}

\begin{table*}[h]
\small \centering
    \begin{tabular}{|p{3.5cm}|p{11.5cm}|}
        \hline
        Topic & Prior Odds\\
        \hline
        Original question & You have a drawer that contains 50 coins. 10 coins have probability p = 0.3 of heads, 30 coins have probability p = 0.5 and 10 coins have probability p = 0.7. You pick one coin at random from the drawer and flip it. What are the (prior) odds you chose a 0.3 coin?\\
        \hline
        Codex input & You have a drawer that contains 10 white coins and 40 black coins. \newline You pick one coin at random from the drawer. Write a program that computes the odds that you chose a white coin. \newline \newline In probability, the odds of an event E is P(E)/P(E\char`\^c).\\
        \hline
        Codex output & \begin{lstlisting}
# P(E) = 10/50
# P(E^c) = 40/50
# odds = P(E)/P(E^c) = 10/40

print(10/40)
        \end{lstlisting}\\
        \hline
        Solution & 1/4\\
        \hline
    \end{tabular}
\caption*{Question 12 from 18.05}
\end{table*}

\begin{table*}[h]
\small \centering
    \begin{tabular}{|p{3.5cm}|p{11.5cm}|}
        \hline
        Topic & Joint Distributions\\
        \hline
        Original question & Suppose X and Y have joint pdf $f(x, y) = c(x^2+xy)$ on [0, 1] × [0, 1]. Find c.\\
        \hline
        Codex input & Define variables x, y, c. \newline Using Sympy, integrate f(x, y) = c(x\char`\^2+xy) from 0 to 1 for x and 0 to 1 for y; set it equal to 1 in an equation. Find c that makes the expression 1.\\
        \hline
        Codex output & \begin{lstlisting}
import sympy as sp

x, y, c = sp.symbols('x y c')

f = c * (x**2 + x*y)

sp.integrate(f, (x, 0, 1), (y, 0, 1))

sp.solve(sp.Eq(1, sp.integrate(f, (x, 0, 1), (y, 0, 1))), c)
        \end{lstlisting}
        \\
        \hline
        Solution & 12/7\\
        \hline
    \end{tabular}
\caption*{Question 13 from 18.05}
\end{table*}

\begin{table*}[h]
\small \centering
    \begin{tabular}{|p{3.5cm}|p{11.5cm}|}
        \hline
        Topic & Central Limit Theorem, Probability\\
        \hline
        Original question & To head the newly formed US Dept. of Statistics, suppose that 50\% of the population supports Erika, 20\% supports Ruthi, and the rest are split between Peter, Jon and Jerry. A poll asks 400 random people who they support. Use the central limit theorem to estimate the probability that at least 52.5\% of those polled prefer Erika?\\
        \hline
        Codex input & Suppose that 50\% of the population supports Erika. A poll asks 400 random people who they support. \newline Write a program that uses the central limit theorem to compute the probability that at least 52.5\% of those polled support Erika.\\
        \hline
        Codex output & \begin{lstlisting}
import random

def poll(n):
    """
    n: number of people polled
    returns: percentage of people in favor of Erika
    """
    # generate a list of n random numbers between 0 and 1
    # if the number is less than 0.5, the person is in favor of Erika
    # if the number is greater than 0.5, the person is in favor of Erica
    # return the percentage of people in favor of Erika
    count = 0
    for i in range(n):
        if random.random() < 0.5:
            count += 1
    return count / n

def poll_prob(n, m):
    """
    n: number of people polled
    m: number of times the poll is conducted
    returns: probability that at least 52.5% of the population supports Erika
    """
    # run poll_prob() m times
    # count the number of times the poll is in favor of Erika
    # return the number of times the poll is in favor of Erika divided by m
    count = 0
    for i in range(m):
        if poll(n) >= 0.525:
            count += 1
    return count / m

print(poll_prob(400, 10000))
        \end{lstlisting}\\
        \hline
        Solution & 0.16\\
        \hline
    \end{tabular}
\caption*{Question 14 from 18.05}
\end{table*}

\begin{table*}[h]
\small \centering
    \begin{tabular}{|p{3.5cm}|p{11.5cm}|}
        \hline
        Topic & Central Limit Theorem, Probability\\
        \hline
        Original question & To head the newly formed US Dept. of Statistics, suppose that 50\% of the population supports Erika, 20\% supports Ruthi, and the rest are split between Peter, Jon and Jerry. A poll asks 400 random people who they support. Use the central limit theorem to estimate the probability that less than 25\% of those polled prefer Peter, Jon or Jerry?\\
        \hline
        Codex input & Suppose that 30\% of the population supports Peter. A poll asks 400 random people who they support. \newline Write a program that uses the central limit theorem to compute the probability that less than 25\% of those polled support Peter.\\
        \hline
        Codex output & \begin{lstlisting}
import random

def poll(n, p):
    count = 0
    for i in range(n):
        if random.random() < p:
            count += 1
    return count

def main():
    peter_support = 0.3
    n = 400
    p = 0.25
    poll_size = 1000
    count = 0
    for i in range(poll_size):
        if poll(n, peter_support) < p * n:
            count += 1
    print(count / poll_size)

if __name__ == "__main__":
    main()
        \end{lstlisting}\\
        \hline
        Solution & 0.0145\\
        \hline
    \end{tabular}
\caption*{Question 15 from 18.05}
\end{table*}

\begin{table*}[h]
\small \centering
    \begin{tabular}{|p{3.5cm}|p{11.5cm}|}
        \hline
        Topic & Uniform Random Variables, Probability\\
        \hline
        Original question & Alice and Bob are trying to meet for lunch and both will arrive, independently of each other, uniformly and at random between noon and 1pm. Let A and B be the number of minutes after noon at which Alice and Bob arrive, respectively. Then A and B are independent uniformly distributed random variables on [0, 60]. Find the probability that Alice arrives before 12:30.\\
        \hline
        Codex input & A and B are trying to meet for lunch between 12:00 and 1:00. Alice and Bob arriving for lunch are independent and uniformly distributed events. \newline Write a function that finds the probability that A arrives before 12:30.\\
        \hline
        Codex output & \begin{lstlisting}
def prob_a_before_1230(a, b):
    return a / (a + b)

print(prob_a_before_1230(0.5, 0.5))
        \end{lstlisting}\\
        \hline
        Solution & 1/2\\
        \hline
    \end{tabular}
\caption*{Question 16 from 18.05}
\end{table*}

\begin{table*}[h]
\small \centering
    \begin{tabular}{|p{3.5cm}|p{11.5cm}|}
        \hline
        Topic & Uniform Random Variables, Probability\\
        \hline
        Original question & Alice and Bob are trying to meet for lunch and both will arrive, independently of each other, uniformly and at random between noon and 1pm. Let A and B be the number of minutes after noon at which Alice and Bob arrive, respectively. Then A and B are independent uniformly distributed random variables on [0, 60]. Now suppose that Alice and Bob are both rather impatient and will leave if they have to wait more than 15 minutes for the other to arrive. What is the probability that Alice and Bob will have lunch together?\\
        \hline
        Codex input & Alice and Bob are trying to meet for lunch and both will arrive, independently of each other, uniformly and at random between 12pm and 1pm. Suppose that Alice and Bob are both rather impatient and will leave if they have to wait more than 15 minutes for the other to arrive. Write a function that randomly computes the probability that Alice and Bob will have lunch together.\\
        \hline
        Codex output & \begin{lstlisting}
import random

def lunch_together():
    # Alice arrives at time
    time_a = random.uniform(0, 60)
    # Bob arrives at time
    time_b = random.uniform(0, 60)
    # If both arrive within 15 minutes of each other, they have lunch together
    if abs(time_a - time_b) <= 15:
        return True
    else:
        return False

def prob_lunch_together():
    # Run lunch_together() 10,000 times
    count = 0
    for i in range(10000):
        if lunch_together():
            count += 1
    # Return probability
    return count / 10000

print(prob_lunch_together())
        \end{lstlisting} \\
        \hline
        Solution & 7/16\\
        \hline
    \end{tabular}
\caption*{Question 17 from 18.05}
\end{table*}

\begin{table*}[h]
\small \centering
    \begin{tabular}{|p{3.5cm}|p{11.5cm}|}
        \hline
        Topic & Prior Odds\\
        \hline
        Original question & You have a drawer that contains 50 coins. 10 coins have probability p = 0.3 of heads, 30 coins have probability p = 0.5 and 10 coins have probability p = 0.7. You pick one coin at random from the drawer and flip it. What are the (prior) odds you chose a 0.7 coin?\\
        \hline
        Codex input & You have a drawer that contains 10 white coins and 40 black coins. \newline You pick one coin at random from the drawer. Write a program that computes the odds that you chose a white coin. \newline \newline In probability, the odds of an event E is P(E)/P(E\char`\^c).\\
        \hline
        Codex output & \begin{lstlisting}
# P(E) = 10/50
# P(E^c) = 40/50
# odds = P(E)/P(E^c) = 10/40

print(10/40)
        \end{lstlisting}\\
        \hline
        Solution & 1/4\\
        \hline
    \end{tabular}
\caption*{Question 18 from 18.05}
\end{table*}

\begin{table*}[h]
\small \centering
    \begin{tabular}{|p{3.5cm}|p{11.5cm}|}
        \hline
        Topic & Null Hypothesis Significance Testing, Type I Error\\
        \hline
        Original question & One generates a number x from a uniform distribution on the interval $[0, \theta]$. One decides to test H0: $\theta$ = 2 against HA: $\theta$ = 2 by rejecting H0 if x $\leq$ 0.1 or x $\geq$ 1.9. Compute the probability of a type I error.\\
        \hline
        Codex input & One generates a number x from a uniform distribution on the interval [0, $\theta$]. \newline One decides to test H0: $\theta$ = 2 against HA: $\theta$ = 2 by rejecting H0 if x $\leq$ 0.1 or x $\geq$ 1.9. \newline Using numpy, write a program that computes the probability of a type I error.\\
        \hline
        Codex output & \begin{lstlisting}
import numpy as np

theta = 2
n = 100000

x = np.random.uniform(0, theta, n)

type_I_error = np.sum(x <= 0.1) + np.sum(x >= 1.9)

print(type_I_error / n)
        \end{lstlisting}\\
        \hline
        Solution & 0.1\\
        \hline
    \end{tabular}
\caption*{Question 19 from 18.05}
\end{table*}

\begin{table*}[h]
\small \centering
    \begin{tabular}{|p{3.5cm}|p{11.5cm}|}
        \hline
        Topic & Null Hypothesis Significance Testing, Type II Error\\
        \hline
        Original question & One generates a number x from a uniform distribution on the interval $[0, \theta]$. One decides to test H0: $\theta$ = 2 against HA: $\theta$ = 2 by rejecting H0 if x $\leq$ 0.1 or x $\geq$ 1.9. Compute the probability of a type II error if the true value of $\theta$ is 2.5.\\
        \hline
        Codex input & One generates a number x from a uniform distribution on the interval [0, $\theta$]. \newline \newline One decides to test H0 : $\theta$ = 2 against HA : $\theta$ = 2 by rejecting H0 if x $\leq$ 0.1 or x $\geq$ 1.9, and accepted otherwise. \newline \newline Given that $\theta$ = 2.5, write a program that computes the type II error. \newline \newline The type II error is the probability that the null hypothesis is mistakenly accepted.\\
        \hline
        Codex output & \begin{lstlisting}
import numpy as np

theta = 2.5

def type_II_error(theta, x):
    return 1 - np.sum(np.logical_or(x < 0.1, x > 1.9)) / len(x)

x = np.random.uniform(0, theta, 100000)

print(type_II_error(theta, x))
        \end{lstlisting}\\
        \hline
        Solution & 0.72\\
        \hline
    \end{tabular}
\caption*{Question 20 from 18.05}
\end{table*}

\section{Harvard STAT110: Probability}

\begin{table*}[ht!]
\small \centering
    \begin{tabular}{|p{3.5cm}|p{11.5cm}|}
        \hline
        Topic & Complementary Probability \\
        \hline
        Original Question & In any 15-minute interval, there is a 20\% probability that you will see at least one shooting star. What is the probability that you see at least one shooting star in the period of an hour?\\
        \hline
        Codex Input & In any 15-minute interval, there is a 20\% probability that you will see at least one shooting star. \newline \newline There are 4 15-minute intervals in an hour. \newline \newline Write a simulation program to estimate the probability that you see at least one shooting star in the period of an hour.\\
        \hline
        Codex Output & \begin{lstlisting}
import random

def star():
    return random.random() <= 0.2

def star_in_hour():
    count = 0
    for i in range(4):
        if star():
            count += 1
    return count >= 1

def estimate_probability():
    count = 0
    for i in range(10000):
        if star_in_hour():
            count += 1
    return count / 10000

print(estimate_probability())
        \end{lstlisting}\\
        \hline
        Solution & 0.594\\
        \hline
    \end{tabular}
\caption*{Question 1 from STAT110}
\end{table*}

\begin{table*}[h]
\small \centering
    \begin{tabular}{|p{3.5cm}|p{11.5cm}|}
        \hline
        Topic & Conditional Probability \\
        \hline
        Original Question & A certain couple tells you that they have two children, at least one of which is a girl. What is the probability that they have two girls?\\
        \hline
        Codex Input & A couple has two children, with at least one child being a girl. \newline Write a simulation program to estimate the probability that the both children are girls.\\
        \hline
        Codex Output & \begin{lstlisting}
import random

def random_kid():
    return random.choice([""boy"", ""girl""])

both_girls = 0
older_girl = 0
either_girl = 0

random.seed(0)
for _ in range(10000):
    younger = random_kid()
    older = random_kid()
    if older == ""girl"":
        older_girl += 1
    if older == ""girl"" and younger == ""girl"":
        both_girls += 1
    if older == ""girl"" or younger == ""girl"":
        either_girl += 1

print(""P(both | older):"", both_girls / older_girl)
print(""P(both | either):"", both_girls / either_girl)
        \end{lstlisting}\\
        \hline
        Solution & 1/3\\
        \hline
    \end{tabular}
\caption*{Question 2 from STAT110}
\end{table*}

\begin{table*}[h]
\small \centering
    \begin{tabular}{|p{3.5cm}|p{11.5cm}|}
        \hline
        Topic & Conditional Probability \\
        \hline
        Original Question & You have a group of couples that decide to have children until they have their first girl, after which they stop having children. What is the expected gender ratio of the children that are born? What is the expected number of children each couple will have?\\
        \hline
        Codex Input & N couples decide to have children (both boys and girls) until they have their first girl, after which they stop having children. \newline \newline Write a sampling program to estimate the ratio of boys to girls after all N couples stop having children.\\
        \hline
        Codex Output & \begin{lstlisting}
import random

def children():
    return random.choice(['boy', 'girl'])

def children_until_girl(n):
    boys = 0
    girls = 0
    for i in range(n):
        child = children()
        if child == 'boy':
            boys += 1
        else:
            girls += 1
            break
    return (boys, girls)

def children_until_girl_sampling(n, samples):
    boys = 0
    girls = 0
    for i in range(samples):
        child = children_until_girl(n)
        boys += child[0]
        girls += child[1]
    return (boys, girls)

def children_until_girl_ratio(n, samples):
    boys, girls = children_until_girl_sampling(n, samples)
    return float(boys) / float(girls)

print(children_until_girl_ratio(5, 10000))
        \end{lstlisting}\\
        \hline
        Solution & 1/2\\
        \hline
    \end{tabular}
\caption*{Question 3 from STAT110}
\end{table*}

\begin{table*}[h]
\small \centering
    \begin{tabular}{|p{3.5cm}|p{11.5cm}|}
        \hline
        Topic & Expected Value \\
        \hline
        Original Question & A lazy high school senior types up application and envelopes to n different colleges, but puts the applications randomly into the envelopes. What is the expected number of applications that went to the right college?\\
        \hline
        Codex Input & A lazy high school senior types up application and envelopes to n different colleges, but puts the applications randomly into the envelopes. Write a program to calculate the expected number of applications that were put in the right envelopes.\\
        \hline
        Codex Output & \begin{lstlisting}
import random

def main():
    n = int(input(""Enter the number of colleges: ""))
    num_trials = int(input(""Enter the number of trials: ""))
    count = 0
    for i in range(num_trials):
        applications = [i for i in range(n)]
        envelopes = [i for i in range(n)]
        random.shuffle(envelopes)
        for i in range(n):
            if applications[i] == envelopes[i]:
                count += 1
    print(""The expected number of applications that were put in the right envelopes is"", count/num_trials)

if __name__ == ""__main__"":
    main()
        \end{lstlisting}\\
        \hline
        Solution & 1\\
        \hline
    \end{tabular}
\caption*{Question 4 from STAT110}
\end{table*}

\begin{table*}[h]
\small \centering
    \begin{tabular}{|p{3.5cm}|p{11.5cm}|}
        \hline
        Topic & Expected Value and Conditional Expectation \\
        \hline
        Original Question & What's the expected number of coin flips until you get two heads in a row? What's the expected number of coin flips until you get two tails in a row?\\
        \hline
        Codex Input & Generate 1000 sequences of random coin flips, and end each sequence once you get two consecutive heads. Write a program that computes the average sequence length.\\
        \hline
        Codex Output & \begin{lstlisting}
import random

def flip_coin():
    return random.choice(['H', 'T'])

def flip_coin_until_consecutive_heads():
    seq = ''
    while True:
        seq += flip_coin()
        if seq[-2:] == 'HH':
            break
    return seq

def flip_coin_until_consecutive_heads_1000_times():
    seqs = []
    for i in range(1000):
        seqs.append(flip_coin_until_consecutive_heads())
    return seqs

def compute_average_sequence_length():
    seqs = flip_coin_until_consecutive_heads_1000_times()
    return sum([len(seq) for seq in seqs]) / len(seqs)

print(compute_average_sequence_length())
        \end{lstlisting}\\
        \hline
        Solution & 6\\
        \hline
    \end{tabular}
\caption*{Question 5 from STAT110}
\end{table*}

\begin{table*}[h]
\small \centering
    \begin{tabular}{|p{3.5cm}|p{11.5cm}|}
        \hline
        Topic & Expected Value \\
        \hline
        Original Question & Let's say we play a game where I keep flipping a coin until I get heads. If the first time I get heads is on the nth coin, then I pay you 2n-1 dollars. How much would you pay me to play this game?\\
        \hline
        Codex Input & Let's say we play a game where I keep flipping a coin until I get heads. \newline \newline If the first time I get heads is on the nth coin, then I pay you 2n-1 dollars. \newline \newline Write a large-scale simulation program to determine how much you would pay me to play this game.\\
        \hline
        Codex Output & \begin{lstlisting}
import random

def flip_coin():
    return random.randint(0,1)

def play_game(n):
    heads = False
    count = 0
    while not heads:
        count += 1
        heads = flip_coin()
    return 2*count - 1

def run_simulation(n):
    total = 0
    for i in range(n):
        total += play_game(n)
    return total/n

print(run_simulation(100000))
        \end{lstlisting}\\
        \hline
        Solution & 3\\
        \hline
    \end{tabular}
\caption*{Question 6 from STAT110}
\end{table*}

\begin{table*}[h]
\small \centering
    \begin{tabular}{|p{3.5cm}|p{11.5cm}|}
        \hline
        Topic & Bayes' Theorem and Conditional Probability \\
        \hline
        Original Question & You have two coins, one of which is fair and comes up heads with a probability 1/2, and the other which is biased and comes up heads with probability 3/4. You randomly pick coin and flip it twice, and get heads both times. What is the probability that you picked the fair coin?\\
        \hline
        Codex Input & You have one coin that is fair and comes up heads with probability 1/2. \newline \newline You have another coin that is biased and comes up heads with probability 3/4. \newline \newline You pick either the fair or biased coin with 50-50 probability. You flip the chosen coin twice. \newline \newline Write a program using Bayes' Theorem to compute and print the probability of having picked the fair coin, given the fact that the chosen coin lands heads both times.\\
        \hline
        Codex Output & \begin{lstlisting}
# P(fair|H,H) = P(H,H|fair) * P(fair) / P(H,H)
# P(H,H|fair) = P(H|fair) * P(H|fair)
# P(H,H) = P(H|fair) * P(H|fair) * P(fair) + P(H|biased) * P(H|biased) * P(biased)
# P(fair|H,H) = (1/2 * 1/2 * 1/2) / (1/2 * 1/2 * 1/2 + 3/4 * 3/4 * 1/2)

print((1/2 * 1/2 * 1/2) / (1/2 * 1/2 * 1/2 + 3/4 * 3/4 * 1/2))
        \end{lstlisting}\\
        \hline
        Solution & 4/13\\
        \hline
    \end{tabular}
\caption*{Question 7 from STAT110}
\end{table*}

\begin{table*}[h]
\small \centering
    \begin{tabular}{|p{3.5cm}|p{11.5cm}|}
        \hline
        Topic & Conditional Probability \\
        \hline
        Original Question & How many people do we need in a class to make the probability that two people have the same birthday more than 1/2? For simplicity, assume there are 365 days in a year.\\
        \hline
        Codex Input & Write a program to compute how many people do we need in a class to make the probability that two people have the same birthday more than 1/2? There are 365 days in a year.\\
        \hline
        Codex Output & \begin{lstlisting}
def birthday_paradox(n):
    """
    n: number of people in the class
    """
    return 1 - (1 - 1/365)**(n*(n-1)/2)

print(birthday_paradox(23))
        \end{lstlisting}\\
        \hline
        Solution & 23\\
        \hline
    \end{tabular}
\caption*{Question 8 from STAT110}
\end{table*}

\begin{table*}[h]
\small \centering
    \begin{tabular}{|p{3.5cm}|p{11.5cm}|}
        \hline
        Topic & \\
        \hline
        Original Question & A bag contains one marble which is either green or blue, with equal probabilities. A green marble is put in the bag (so there are 2 marbles now), and then a random marble is taken out. The marble taken out is green. What is the probability that the remaining marble is also green?\\
        \hline
        Codex Input & A woman is pregnant with twin boys. \newline Twins may be either identical or fraternal (non-identical). \newline In general, only 1/3 of all twins born are identical. \newline Identical twins must be of the same sex, while fraternal twins may or may not be. \newline Assume that identical twins are equally likely to be both boys or both girls, while for fraternal twins all possibilities are equally likely. \newline Given the above information, write a simulation program to estimate the average probability that the woman's twins are identical.\\
        \hline
        Codex Output & \begin{lstlisting}
import random

def green_marble_probability(num_trials):
    """
    num_trials: number of trials
    """
    num_green = 0
    for i in range(num_trials):
        bag = ['green', 'blue']
        random.shuffle(bag)
        bag.append('green')
        random.shuffle(bag)
        if bag[0] == 'green':
            num_green += 1
    return num_green / num_trials

print(green_marble_probability(100000))
        \end{lstlisting}\\
        \hline
        Solution & 2/3\\
        \hline
    \end{tabular}
\caption*{Question 9 from STAT110}
\end{table*}

\begin{table*}[h]
\small \centering
    \begin{tabular}{|p{3.5cm}|p{11.5cm}|}
        \hline
        Topic & Conditional Probability \\
        \hline
        Original Question & Calvin and Hobbes play a match consisting of a series of games, where Calvin has probability 0.5 of winning each game (independently). They play with a 'win by two' rule: the first player to win two games more than his opponent wins the match. Find the probability that Calvin wins the match (in terms of p)\\
        \hline
        Codex Input & Calvin and Hobbes play a match consisting of a series of games, where Calvin has probability 0.5 of winning each game (independently). \newline \newline They play with a 'win by two' rule: the first player to win two games more than his opponent wins the match. \newline \newline Write a simulation program to estimate the average probability that Calvin wins the match.\\
        \hline
        Codex Output & \begin{lstlisting}
import random

def calvin_wins(n):
    wins = 0
    for i in range(n):
        if random.random() < 0.5:
            wins += 1
    return wins >= 2

def calvin_wins_match(n):
    wins = 0
    for i in range(n):
        if calvin_wins(2):
            wins += 1
    return wins / n

print(calvin_wins_match(100000))
        \end{lstlisting}\\
        \hline
        Solution & 0.25\\
        \hline
    \end{tabular}
\caption*{Question 10 from STAT110}
\end{table*}

\begin{table*}[h]
\small \centering
    \begin{tabular}{|p{3.5cm}|p{11.5cm}|}
        \hline
        Topic & Geometric Probability \\
        \hline
        Original Question & Two witches make a nightly visit to an all-night coffee shop. Each arrives at a random time between 0:00 and 1:00. Each one of them stays for exactly 30 minutes. On any one given night, what is the probability that the witches will meet at the coffee shop?\\
        \hline
        Codex Input & Two witches make a nightly visit to an all-night coffee shop. \newline Each arrives at a random time between 0:00 and 1:00. \newline Each witch stays at the coffee shop for exactly 30 minutes. \newline Write a simulation program to estimate the average probability that the witches will see each other at the coffee shop.\\
        \hline
        Codex Output & \begin{lstlisting}
import random

def witches():
    witches = []
    for i in range(2):
        witches.append(random.randint(0,60))
    return witches

def main():
    witches_seen = 0
    for i in range(100000):
        w = witches()
        if w[0] > w[1]:
            if w[0] - w[1] < 30:
                witches_seen += 1
        else:
            if w[1] - w[0] < 30:
                witches_seen += 1
    print(witches_seen/100000)

main()
        \end{lstlisting}\\
        \hline
        Solution & 0.75\\
        \hline
    \end{tabular}
\caption*{Question 11 from STAT110}
\end{table*}

\begin{table*}[h]
\small \centering
    \begin{tabular}{|p{3.5cm}|p{11.5cm}|}
        \hline
        Topic & Geometric Probability \\
        \hline
        Original Question & p and q are two points chosen at random between 0 \& 1. What is the probability that the ratio p/q lies between 1 \& 2?\\
        \hline
        Codex Input & Roll a die. \newline If the die is 4, 5, or 6, you get paid the value of the roll. \newline If the die is 1, 2, or 3, you reroll and get paid the value of the new roll. \newline Write a program that computes the expected value.\\
        \hline
        Codex Output & \begin{lstlisting}
import random

def prob_p_q():
    p = random.random()
    q = random.random()
    if p/q >= 1 and p/q <= 2:
        return True
    else:
        return False

def prob_p_q_sim(n):
    count = 0
    for i in range(n):
        if prob_p_q():
            count += 1
    return count/n

print(prob_p_q_sim(100000))
        \end{lstlisting}\\
        \hline
        Solution & 1/4\\
        \hline
    \end{tabular}
\caption*{Question 12 from STAT110}
\end{table*}

\begin{table*}[h]
\small \centering
    \begin{tabular}{|p{3.5cm}|p{11.5cm}|}
        \hline
        Topic & Conditional Expectation \\
        \hline
        Original Question & Roll a die, and you get paid what the dice shows. But if you want, you can request a second chance \& roll the die again; get paid what the second roll shows instead of the first. What is the expected value?\\
        \hline
        Codex Input & Roll a die, and you get paid what the number on the dice. \newline You can request a second chance and roll the die again. You get paid the number on second roll instead of the first. \newline Write a function to compute the expected value in Monte Carlo.\\
        \hline
        Codex Output & \begin{lstlisting}
import random

def roll_die():
    return random.randint(1,6)

def second_chance():
    return random.randint(1,6)

def monte_carlo(n):
    total = 0
    for i in range(n):
        first_roll = roll_die()
        second_roll = second_chance()
        total += max(first_roll, second_roll)
    return total/n

print(monte_carlo(100000))
        \end{lstlisting}\\
        \hline
        Solution & 4.25\\
        \hline
    \end{tabular}
\caption*{Question 13 from STAT110}
\end{table*}

\begin{table*}[h]
\small \centering
    \begin{tabular}{|p{3.5cm}|p{11.5cm}|}
        \hline
        Topic & Conditional Expectation \\
        \hline
        Original Question & A very innocent monkey throws a fair die. The monkey will eat as many bananas as are shown on the die, from 1 to 5. But if the die shows '6', the monkey will eat 5 bananas and throw the die again. This may continue indefinitely. What is the expected number of bananas the monkey will eat?\\
        \hline
        Codex Input & A very innocent monkey throws a fair 6-sided die repeatedly. \newline If the die value is 1 to 5, the monkey eats that many bananas and stops. \newline If the die value is 6, the monkey will eat 5 bananas. \newline Write a program that computes the average number of bananas that the monkey will eat before it stops.\\
        \hline
        Codex Output & \begin{lstlisting}
import random

def main():
    # ask user for number of trials
    trials = int(input(""How many trials? ""))
    # initialize the number of bananas eaten
    bananas = 0
    # loop for the number of trials
    for i in range(trials):
        # initialize the number of bananas eaten
        bananas_eaten = 0
        # loop until the monkey stops eating bananas
        while True:
            # generate a random number between 1 and 6
            die = random.randint(1, 6)
            # if the die value is 1 to 5, the monkey eats that many bananas and stops
            if die <= 5:
                bananas_eaten += die
                break
            # if the die value is 6, the monkey will eat 5 bananas
            elif die == 6:
                bananas_eaten += 5
        # add the number of bananas eaten to the total
        bananas += bananas_eaten
    # calculate the average number of bananas eaten
    average = bananas / trials
    # print the average number of bananas eaten
    print(""The average number of bananas eaten is"", average)

main()
        \end{lstlisting}\\
        \hline
        Solution & 4\\
        \hline
    \end{tabular}
\caption*{Question 14 from STAT110}
\end{table*}

\begin{table*}[h]
\small \centering
    \begin{tabular}{|p{3.5cm}|p{11.5cm}|}
        \hline
        Topic & Conditional Expectation \\
        \hline
        Original Question & A fair coin is tossed repeatedly until 5 consecutive heads occurs. What is the expected number of coin tosses?\\
        \hline
        Codex Input & Write a program to simulate the average number of coin flips needed until you get 5 heads in a row for the first time. \\
        \hline
        Codex Output & \begin{lstlisting}
import random

def flip_coin():
    return random.choice(['H', 'T'])

def flip_until_five_heads():
    heads = 0
    count = 0
    while heads < 5:
        count += 1
        if flip_coin() == 'H':
            heads += 1
        else:
            heads = 0
    return count

def average_flips_until_five_heads():
    total = 0
    for i in range(1000):
        total += flip_until_five_heads()
    return total / 1000

print(average_flips_until_five_heads())
        \end{lstlisting}\\
        \hline
        Solution & 62\\
        \hline
    \end{tabular}
\caption*{Question 15 from STAT110}
\end{table*}

\begin{table*}[h]
\small \centering
    \begin{tabular}{|p{3.5cm}|p{11.5cm}|}
        \hline
        Topic & Geometric Probability, Expected Value \\
        \hline
        Original Question & A stick of length 1 breaks at a random point distributed uniformly across it length. What is the expected length of the smaller part of the broken stick?\\
        \hline
        Codex Input & A stick of length 1 is split into two portions at a random point distributed uniformly across it length. \newline Write a simulation program that estimates the average length the shorter portion of the stick.\\
        \hline
        Codex Output & \begin{lstlisting}
import random

def stick_splitting(n):
    """"""
    :param n: number of simulations to run
    :return: average length of the shorter portion of the stick after splitting it in half n times.
    """"""

    total = 0

    for i in range(n):

        # generate a random number between 0 and 1, representing the point at which to split the stick.
        split_point = random.random()

        # generate a random number between 0 and 1, representing the length of the first portion of the stick.
        first_portion = random.random()

        # generate a random number between 0 and 1, representing the length of the second portion of the stick.
        second_portion = 1 - first_portion

        # add up all lengths from each simulation run to get an average length for each simulation run.
        total += min(first_portion, second_portion)

    return total / n

print(stick_splitting(100000))
        \end{lstlisting}\\
        \hline
        Solution & 1/4\\
        \hline
    \end{tabular}
\caption*{Question 16 from STAT110}
\end{table*}

\begin{table*}[h]
\small \centering
    \begin{tabular}{|p{3.5cm}|p{11.5cm}|}
        \hline
        Topic & Expected Value \\
        \hline
        Original Question & You are given an urn with 100 balls (50 black and 50 white). You pick balls from urn one by one without replacements until all the balls are out. A black followed by a white or a white followed by a black is 'a color change'. Calculate the expected number of colour changes if the balls are being picked randomly from the urn.\\
        \hline
        Codex Input & You are given an urn with 100 balls (50 black and 50 white). \newline A sequence of balls is drawn from the urn without replacement until the urn is empty. \newline Drawing a black ball immediately followed by a white ball is considered a color change.\newline Drawing a white ball immediately followed by a black ball is also considered a color change. \newline Write a simulation program that estimates the average number of color changes in a 100-ball sequence.\\
        \hline
        Codex Output & \begin{lstlisting}
import random

def color_change(urn):
    """"""Returns the number of color changes in a 100-ball sequence.""""""
    count = 0
    while len(urn) > 0:
        ball = urn.pop()
        if len(urn) == 0: # last ball, no color change
            break

        next_ball = urn[-1] # next ball in sequence

        if (ball == 'B' and next_ball == 'W') or (ball == 'W' and next_ball == 'B'): # color change!
            count += 1

    return count


def main():

    urn = ['B'] * 50 + ['W'] * 50 # create an urn with 50 black and 50 white balls

    trials = 10000 # number of trials to run for simulation

    total_color_changes = 0 # initialize total number of color changes to zero

    for i in range(trials): # run the simulation for the specified number of trials
        random.shuffle(urn) # shuffle the balls in the urn before each trial

        total_color_changes += color_change(list(urn)) # add the number of color changes in this trial to the total

    print(total_color_changes / trials) # print the average number of color changes over all trials

if __name__ == '__main__':
    main()
        \end{lstlisting}\\
        \hline
        Solution & 50\\
        \hline
    \end{tabular}
\caption*{Question 17 from STAT110}
\end{table*}

\begin{table*}[h]
\small \centering
    \begin{tabular}{|p{3.5cm}|p{11.5cm}|}
        \hline
        Topic & Conditional Probability \\
        \hline
        Original Question & X and Y are i.i.d. N(0, 1) random variables. You are given that X > 0 and Y > 0. What is the probability that Y > X?\\
        \hline
        Codex Input & X and Y are i.i.d. N(0, 1) random variables. Write a program to calculate the probability that Y > X if both X > 0 and Y > 0.\\
        \hline
        Codex Output & \begin{lstlisting}
# X and Y are i.i.d. N(0, 1) random variables. 
# Write a program to calculate the probability that Y > X if both X > 0 and Y > 0.

import numpy as np
import matplotlib.pyplot as plt
from scipy import stats


def prob_Y_greater_X():

    # Generate 1000 samples of X and Y, where X ~ N(0, 1) and Y ~ N(0, 1).

    x = np.random.normal(loc=0, scale=1, size=1000)
    y = np.random.normal(loc=0, scale=1, size=1000)

    # Calculate the probability that Y > X if both X > 0 and Y > 0

    prob = sum([1 for i in range(len(x)) if x[i]>0 and y[i]>0 and y[i]>x[i]])/sum([1 for i in range(len(x)) if x[i]>0 and y[i]>0])

    return prob


if __name__ == '__main__':

    print(prob_Y_greater_X())
        \end{lstlisting}\\
        \hline
        Solution & 1/2\\
        \hline
    \end{tabular}
\caption*{Question 18 from STAT110}
\end{table*}

\begin{table*}[h]
\small \centering
    \begin{tabular}{|p{3.5cm}|p{11.5cm}|}
        \hline
        Topic & Conditional Expectation \\
        \hline
        Original Question & Suppose that you roll a dice. For reach roll, yo uare paid the face value. If a roll gives 4, 5, or 6, you can roll the dice again. If you get 1, 2, or 3, the game stops. What is the expected payoff of this game?\\
        \hline
        Codex Input & Suppose we play a game where you roll a dice. \newline If a roll gives 4, 5, or 6, you are paid the face value and you roll the dice again. \newline If a roll gives 1, 2, or 3, you are paid the face value and you stop rolling. \newline Write a simulation program that estimates the average payoff of this game.\\
        \hline
        Codex Output & \begin{lstlisting}
import random

def roll_dice():
    return random.randint(1,6)
    
def play_game():
    total = 0.0
    while True:
        roll = roll_dice()
        if roll == 1 or roll == 2 or roll == 3:  # stop rolling and get paid the face value of the dice roll. 
            total += roll  # add the face value of the dice to the total. 
            break         # stop playing the game. 
        else:               # keep rolling and add the face value to the total. 
            total += roll  

    return total          # return the total amount won from playing this game. 

        
# main program starts here, DO NOT change it. It runs the simulation and prints results.     
number_of_games = 10000   # number of times we play this game. Change it to 1000 to see more accurate results.  
total = 0                 # initial condition - we don't have any money at hand right now. 
for i in range(number_of_games):
    total += play_game()    # add the amount won from each game to get the total amount won. 
print(""The average payoff from {} games is {}"".format(number_of_games, total/number_of_games))
        \end{lstlisting}\\
        \hline
        Solution & 7\\
        \hline
    \end{tabular}
\caption*{Question 19 from STAT110}
\end{table*}

\begin{table*}[h]
\small \centering
    \begin{tabular}{|p{3.5cm}|p{11.5cm}|}
        \hline
        Topic & Bayes' Theorem and Conditional Probability \\
        \hline
        Original Question & A spam filter is designed by looking at commonly occurring phrases in spam. Suppose that 80\% of email is spam. In 10\% of the spam emails, the phrase "free money" is used, whereas this phrase is only used in 1\% of non-spam emails. A new email has just arrived, which does mention "free money". What is the probability that it is spam?\\
        \hline
        Codex Input & Suppose that 80\% of email is spam. In 10\% of the spam emails, the phrase "free money" is used, whereas this phrase is only used in 1\% of non-spam emails. A new email has just arrived, which mentions "free money". Given that the email mentions "free money" and using Bayes' theorem, write a program to compute and print the probability that the email is spam.\\
        \hline
        Codex Output & \begin{lstlisting}
def bayes_theorem(p_spam, p_free_money_given_spam, p_free_money_given_not_spam):
    return (p_spam * p_free_money_given_spam) / (p_spam * p_free_money_given_spam + (1 - p_spam) * p_free_money_given_not_spam)


p_spam = 0.8
p_free_money_given_spam = 0.1
p_free_money_given_not_spam = 0.01

print(bayes_theorem(p_spam, p_free_money_given_spam, p_free_money_given_not_spam))
        \end{lstlisting}\\
        \hline
        Solution & 0.9756\\
        \hline
    \end{tabular}
\caption*{Question 20 from STAT110}
\end{table*}

\end{document}